\documentclass{article}

\usepackage{microtype}
\usepackage{graphicx}
\usepackage{subfigure}
\usepackage{booktabs} % for professional tables

\usepackage{hyperref}

\usepackage[accepted]{hill2019}

\icmltitlerunning{The Practical Challenges of Active Learning}

\begin{document}

\twocolumn[
\icmltitle{The Practical Challenges of Active Learning:\\Lessons Learned from Live Experimentation}



\icmlsetsymbol{equal}{*}

\begin{icmlauthorlist}
\icmlauthor{Jean-Fran\c{c}ois Kagy}{google}
\icmlauthor{Tolga Kayadelen}{google}
\icmlauthor{Ji Ma}{google}
\icmlauthor{Afshin Rostamizadeh}{google}
\icmlauthor{Jana Strnadova}{google}
\end{icmlauthorlist}

\icmlaffiliation{google}{Google AI, New York, New York}

\icmlcorrespondingauthor{Jean-Fran\c{c}ois Kagy}{jfkagy@google.com}

\icmlkeywords{Active Learning}

\vskip 0.3in
]



\printAffiliationsAndNotice{}  % leave blank if no need to mention equal contribution
%\printAffiliationsAndNotice{ion} % otherwise use the standard text.

\begin{abstract}
We tested in a live setting the use of active learning for selecting text sentences for human annotations used in training a Thai segmentation machine learning model. In our study, two concurrent annotated samples were constructed, one through random sampling of sentences from a text corpus, and the other through model-based scoring and ranking of sentences from the same corpus. In the course of the experiment, we observed the effect of significant changes to the learning environment which are likely to occur in real-world learning tasks. We describe how our active learning strategy interacted with these events and discuss other practical challenges encountered in using active learning in the live setting.
\end{abstract}

\section{Introduction}
On many supervised tasks, the cost of collecting and annotating data can be reduced by carefully selecting which examples should be sampled and labeled. Diverse active sampling strategies have been proposed to train better models using fewer labeled examples on a wide range of applications; see \citet{settles2009} and the references therein.

Offline (or simulated) experimentation has become the \emph{de facto} standard approach for comparing alternative sampling methods, and this especially holds true for tests of active learning found in academic studies. In the context of evaluating a sampling strategy, an offline experiment consists of using an existing pool of labeled data as a proxy for the population of unlabeled examples (i.e., the sampling pool for the actual annotation task). Samples are drawn from this labeled offline pool using the candidate strategy and, since the labels for the selections are already provided, the performance gains or labeling cost reduction afforded by the sampling strategy can be readily estimated by training models on the simulated samples. Because they reuse preexisting labeled data, offline experiments can be implemented at zero annotation cost. In contrast, in an online (or live) experiment where selections are made directly from the target population of unlabeled examples, the full annotation cost must be incurred as the labels for the selections must be gathered. In addition, offline experiments are easy to design and parallelize, and can be easily reproduced. When tested in such controlled settings, active learning has been shown to largely outperform passive sampling on many classification tasks and datasets.

In spite of its apparent versatility, offline experimentation comes with important limitations for performance benchmarking. Foremost, offline datasets may poorly approximate the target population distribution. Data filtering and cleaning operations, as well as biased sampling methods (e.g., oversampling of the minority class), are commonly used in building training datasets. Furthermore, the use of offline datasets assumes a fixed distribution of ground truth labels, but in reality ``true'' labels are prone to shifts over time. Assuming away the occurrence of these label shifts can lead to false assessments of learning performance. For example, in the live study of active learning by \citet{Baldridge2009HowWD}, the expert linguist was found to score a lower labeling accuracy rate than the non-expert rater, only as a result of the expert's reliance on revised labeling rules which the study was unaware of. Also, as these authors note, in offline experiments, since the simulated samples are typically drawn from a small pool of labeled data, a ``throttling'' effect can occur as the selected training sample exhausts the entire labeled pool and all sampling methods tend to select largely overlapping samples. Such a throttling effect can induce biased comparisons of learning curves.

In light of these limitations, we ought to ask whether active learning effectively works in live environments. While active learning has been previously analyzed in the context of live data collection, e.g., in \citet{Baldridge2009HowWD} and \citet{Druck:2009:ALL:1699510.1699522}, it is our understanding that these studies were run in controlled settings where the learning environment remains unchanged. In this paper, we present a live application of active learning for selecting text sentences for human annotations used in training a Thai segmentation machine learning model. In our study, two concurrent annotated samples were constructed, one through random sampling of sentences from a large unlabeled text corpus, and the other through model-based scoring and ranking of sentences from the same corpus. In the course of the experiment, we observed the effect of significant changes to the learning environment which are likely to occur in real-world learning tasks. These changes include a switch to a new model and a major revision of existing labels. We describe how our active learning strategy interacted with these changes and discuss other practical challenges encountered in using active learning in the live setting. We then propose guidelines addressing each of these challenges which can serve for the design of live experimentation of active learning, and more generally for the application of active learning in live settings.

The outline of this paper is as follows. In Section \ref{design}, we describe the design of our live experiment and present results. In Section \ref{challenges}, we expand on the practical challenges encountered in applying active learning in this experiment. Section \ref{conclusion} concludes.

\section{Live Experiment}\label{design}
Two concurrent samples of text sentences were annotated in the course of the experiment. The experiment had an annotation budget of 30,000 sentences, so each sample was given a target size of 15,000 sentences. The sampled sentences would be sent to human raters who would attach binary labels (``break'', ``no break'') to each token of the sentence (or more precisely, to each Unicode code point). Both samples were selected from the same pool of 1.4 million unannotated sentences. The first sample (\emph{passive}) was uniformly randomly selected from the pool. Under the other scheme (\emph{active}), the sample was selected using margin sampling \citep{Scheffer:2001:AHM:647967.741626}---a particular sampling algorithm in the general family of uncertainty-based methods \citep{settles2009}. To construct an annotated sample using margin sampling, the label probabilities predicted by some chosen model (\emph{scorer}) are used to compute the margin score of each candidate unlabeled example. The candidates having the lowest margin scores are then selected for annotation. Denoting the unlabeled pool by $X$ and the model's predicted label probabilities for sentence $x\in X$ by $p_1(x), p_2(x), ..., p_K(x)$, where the probabilities for the $K$ alternative predictions are assumed to be sorted in decreasing order, the margin score for $x$ is computed as $M(x) = p_1(x) - p_2(x)$. A target sample size of, say $N$ sentences, can be achieved by ranking $\{M(x)|x\in X\}$ in ascending order and selecting the corresponding first $N$ sentences. We applied a token-size penalty as a multiplicative term to avoid selections biased towards the longest sentences.\footnote{Label probabilities generally decrease with sentence length.} The penalized margin score is given as $\tilde{M}(x) = M(x) \cdot L(x)^\lambda$, where $L(x)$ denotes the number of tokens in sentence $x$ and $\lambda$ is a parameter which can be used for tuning the extent of sentence length regularization.

Under both schemes, the selected samples were annotated in batches of 1,000 sentences. We fully re-trained the scorer on the incremental \emph{active} labels after each iteration. Both the \emph{active} and \emph{passive} evaluation models used a common seed of 8,745 sentences among their training data. A common test set of 7,371 sentences was used to compare the performance of the two samples.

During a potentially lengthy data collection and annotation process, we can expect the learning model type/architecture to evolve.
%While the DNN is able to express more complex mappings between the features and the segmentation labels, the perceptron presents many advantages as a scorer. First, by using beam search on the perceptron, we were able to generate the alternative predictions for each example which are required for margin computation. In contrast, because the production DNN segmenter used a greedy solution, only single scores for the top predicted labels were returned by this model. Secondly, in contrast to the DNN which computed a vector of label probabilities---one probability score for each token in the sentence---the perceptron returned sentence-level confidence scores which could directly be used in the margin formula. We thus averted the need of finding an appropriate mapping from token-level scores into a sentence-level score. 
Our experiment setup considered two classes of segmenters: perceptron models and feedforward deep neural network-based (DNN) models.
During the live collection, the perceptron segmenter is the incumbent model and a switch to the DNN segmenter occurs halfway through the experiment (i.e., after eight sampling iterations). Also during the experiment, we observed the effects of a large-scale switching of existing labels akin to a major revision of annotation guidelines. This resulted in the shifting of one or more of the labels for 40\% of the already-annotated sentences (i.e., from ``no break'' to ``break'' labels or vice versa). All of the label switches occurred at the same time, concurrently with the model switch.

\begin{figure*}
    \centering
    \begin{tabular}{cc}
    \includegraphics[scale=0.28]{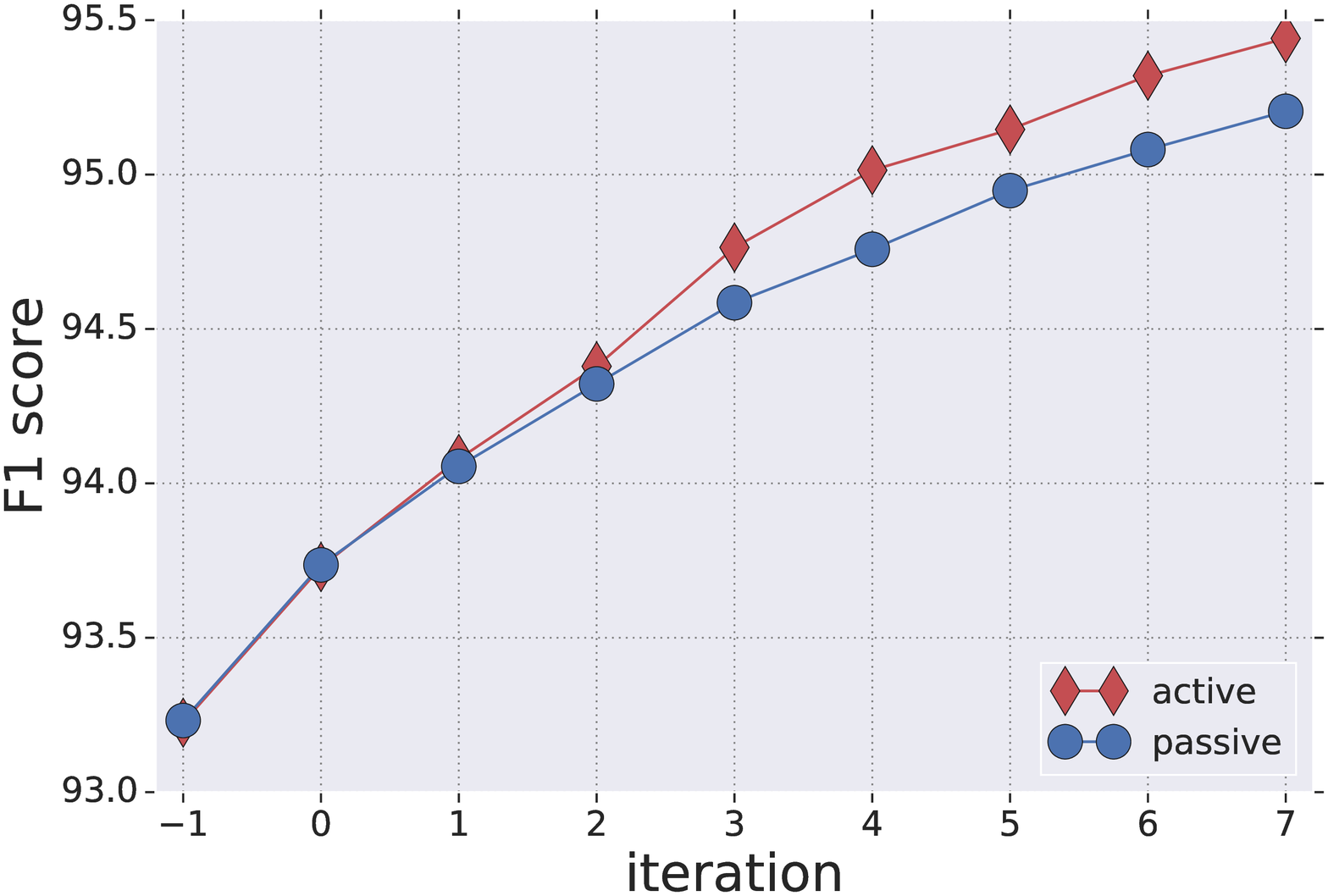} & 
    \includegraphics[scale=0.28]{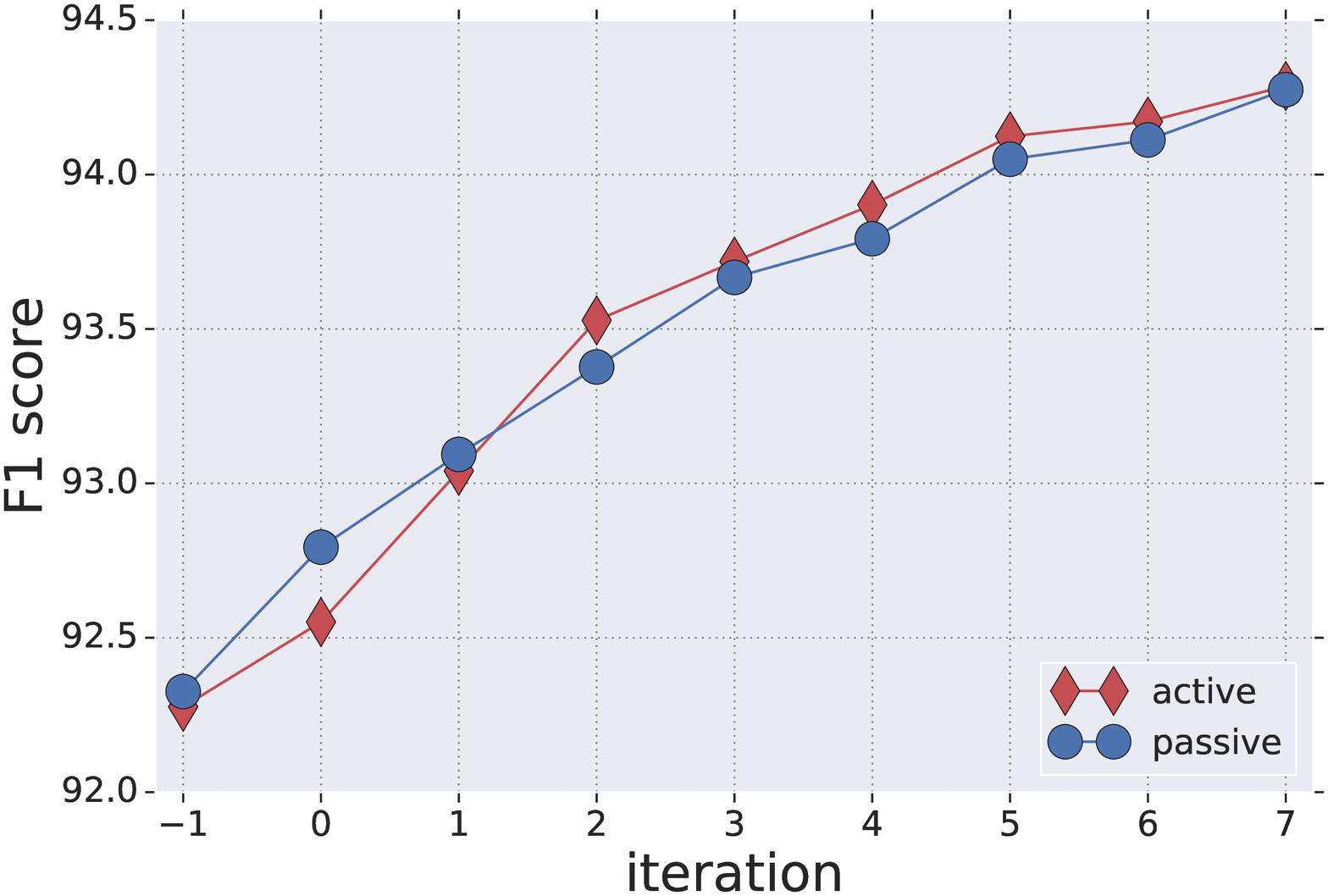} \\
    (a) & (b)
    \end{tabular}
%    \vspace{-0.3cm}
    \caption{
(a): F1 scores of perceptron segmenters trained on the \emph{active} and \emph{passive} samples in the pre-switch iterations.
(b): F1 scores of DNN segmenters trained on the \emph{active} and \emph{passive} samples in the pre-switch iterations. $t=-1$ corresponds to the models trained on the common (pre-switch) seed training set.}
    \label{fig:fig1}
\end{figure*}

Figure~\ref{fig:fig1}(a) compares the F1 score improvements through iteration $t=7$ of perceptron segmenters trained on the \emph{active} and \emph{passive} selections. It indicates that the actively sampled training set can lead to a better model compared to the model trained on a passive sample. Figure~\ref{fig:fig1}(b), which compares the F1 scores of DNN segmenters trained on the same samples, does not showcase the same gains from the \emph{active} samples. From these results, we draw contrasting implications for the effectiveness of active learning. Among others, these figures raise the possibility that the perceptron may poorly approximate which regions of the feature space could be most informative from the standpoint of the DNN segmenter. This hypothesis is supported by findings in \citet{baldridge:osborne} to the effect that lack of relatedness between the scoring and test models can limit the gains from active learning. We will further expand on this issue in Subsection \ref{model_switch_subsection}.

\begin{figure}[t]
    \centering
    \includegraphics[scale=0.28]{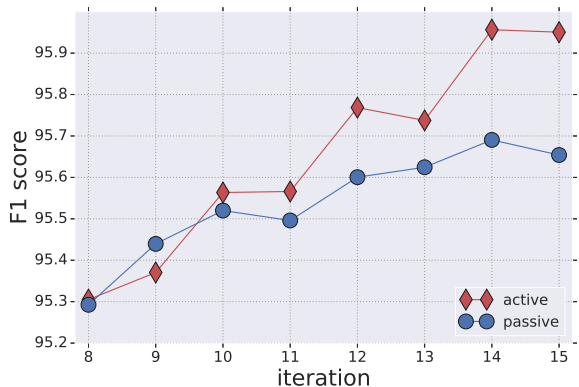}
%    \vspace{-0.5cm}
    \caption{F1 scores of DNN segmenters trained on the \emph{active} and \emph{passive} samples in the post-label revision iterations. $t=8$ corresponds to the models trained on the common (post-switch) seed training set.}
    \label{fig:fig2}
\end{figure}

Following the switch, we used the DNN segmenter as scorer. To measure the performance of active learning in the new DNN regime, we incremented the common training seed set with all of the labels collected through iteration $t=7$. Also, a large set of labels collected outside this experiment had just been provided to us which we used to increment our evaluation set to a total of 52,000 sentences. 
From $t=8$ onwards, new \emph{active} and \emph{passive} samples were collected and evaluated in the same fashion described above. Figure~\ref{fig:fig2} compares the performance of the two samples for $t=8$ onwards. The sampling gains of \emph{active} over \emph{passive} over the post-switch regime roughly amounted to between 33\% to 43\%, meaning the \emph{active} model was able to achieve the same F1 score as the \emph{passive} model using 33\% to 43\% fewer training examples.\footnote{Sampling gains can be defined in different ways. A conservative measurement would be based on the latest iteration at which the \emph{active} model just surpassed the highest F1 score on the \emph{passive} curve (``last-vs-max''). The \emph{active} model used 2,000 fewer training sentences to outperform the best F1 score of the \emph{passive} model which was realized on 6,000 sentences, leading to a last-vs-max gain of 33\%.  A more lenient measurement (``first-vs-final'') would consider the earliest iteration at which the \emph{active} model just surpassed the final \emph{passive} F1 score. The \emph{active} model was able to use as few as 4,000 training sentences to outperform the F1 score of the \emph{passive} model trained on the full \emph{passive} sample of 7,000 sentences, leading to a first-vs-final gain of $3,000 / 7,000 \approx 43\%$.}

\section{Practical Challenges}\label{challenges}
Next, we describe how active learning interacted with the types of changes observed during our experiment and discuss other practical challenges encountered in using active learning in the live setting.

\subsection{Shifting Models}\label{model_switch_subsection}
Figures~\ref{fig:fig1}(b) and \ref{fig:fig2} draw contrasting implications for the effectiveness of active learning. Given the several changes that occurred halfway through the experiment, it would be difficult to pinpoint a single root cause for this difference. We note that among these changes, in the pre-switch regime, the active selections were based on the perceptron as scorer, while in the post-switch regime the DNN was used as scorer. Formally establishing a causal link between model mismatch and the performance of active learning in our live experiment is beyond the scope of this paper. As we discussed in Section \ref{design}, such a link has been investigated in \citet{baldridge:osborne}, albeit in an offline setting. 

In ways related to the impact of model mismatch, our experiment has underscored a corollary question about active learning: can robust sampling methods be applied when models are bound to evolve? This is an important question, as uncertainty about future models can complicate the current decision-making for an appropriate scorer. A number of studies on active learning under model selection (ALMS) have explored related setups where the evaluation model is not fixed but is instead chosen in conjunction with the scoring model. These setups abstract from any model uncertainty: models are treated as pure choice variables. However, since the notion of yet-to-be-determined models is central to these papers, they may be regarded as natural starting points for our own investigation.

In one such study, \citet{ali2014} explores ALMS on classification tasks in the streaming setting. They propose building a concurrent model bias-free validation set separately from the active training set for iteratively learning a posterior distribution of model ranking. Their approach treats the model selection problem as a one-shot decision, where the learned distribution is over the best static model. The superior performance of their ALMS algorithm over an oracle (i.e., active learner who knows the best model in hindsight) suggests that powerful but complex interactions can exist between the optimal choice of training data and the optimal choice of model, even in such a stylized setting. In \citet{LuBongard2009}, active learning is applied on top of an evolutionary algorithm in order to learn the optimal weights of an ensemble of models belonging to a given set of model classes.

Fundamental insights into ALMS can also be drawn from the vast literature on optimal experimental design, a general class of problems of which ALMS could be viewed as a special instance. For example, \citet{sugiyama2008} propose a two-stage ensemble design in the regression setting, where an initial active batch of examples is chosen to minimize a model-weighted expected generalization error. Upon labeling the selected set, the best model is chosen among the candidate models trained on this set. Their ALMS strategy is seen to outperform the sequential (naive) approach in which the active selection is based on the best candidate model among all models trained on the current active set. Again, we see the importance of solving the active learning and model selection problems jointly rather than separately. For other treatments of ALMS, see e.g., \citet{DBLP:journals/corr/abs-1207-4138} and \citet{KapoorRL}.

Given the availability of robust ALMS methods, we ought to ask whether model switches have become non-issues for active learning. First, we note that the outperformance of the few proposed ALMS methods have been shown for only a few datasets and tasks and only in the offline setting. There is little evidence that these gains could be replicated on a live data collection task, as in our experiment. Furthermore, the studies have assumed a static set of models for selection. In practice the notion of model selection set is a dynamic one: research advances bring about new model architectures in a periodic fashion. Having little visibility into the next generation of models, there is not much that we can say on how the training data “tuned” for one particular class of models will interact with these future models.

\begin{figure}
    \centering
    \includegraphics[scale=0.28]{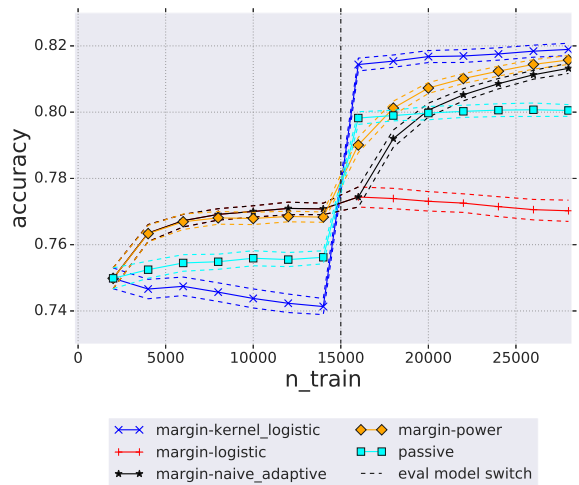}
    \caption{Comparison of sampling schemes on the \emph{covtype.binary} dataset. The prevailing model is assumed to be \emph{logistic} in the initial sampling rounds. After seven rounds, a deterministic switch to \emph{kernel\_logistic} takes place. Solid lines represent the mean test accuracy over 15 trials. Dashed lines represent (approximate) 95\% confidence intervals for the estimated mean accuracy.}
    \label{fig:fig3}
\end{figure}

Aside from shifting model choice sets, the rapid and unpredictable evolution of models also raises other practical questions for active learning. To help us further explore the problem, we ran a simulated experiment on the public \emph{covtype.binary} dataset.\footnote{\url{https://www.csie.ntu.edu.tw/~cjlin/libsvmtools/datasets/binary.html}.} The results of this experiment is shown in Figure~\ref{fig:fig3}. In this stylized setup, the model set consists of a logistic classifier (\emph{logistic}, i.e., the simpler model) and an approximate RBF logistic classifier (\emph{kernel\_logistic}, i.e., the more complex model). On each trial, the dataset is randomly split into a (pseudo) unlabeled pool of 571,012 examples and a holdout set of 10,000 examples used for model evaluation. At each round, a batch of 2,000 examples is selected under the proposed scheme and added to the training set. The prevailing test model is assumed to be \emph{logistic} in the initial rounds. After seven rounds, a deterministic switch to \emph{kernel\_logistic} took place. All of the active learning schemes considered use margin sampling (\emph{margin}) for selecting candidate examples.\footnote{At each round, a subset of 40,000 examples is first randomly selected from the remaining unlabeled pool. Margin sampling is then performed on the subsetted pool.}

The active schemes differ in the model they used for scoring examples. \emph{margin-logistic} and \emph{margin-kernel\_logistic} denote pure schemes, in the sense that the scoring model is from the same class in all rounds. Under \emph{margin-naive\_adaptive}, the current test model is used as scorer on the next batch selection. In the \emph{margin-power} scheme, the scorer is a weighted ensemble of the trained models from each class, where the weight schedule follows a piecewise power law function of the training sample size. This schedule is chosen to optimize the average F1 score over the sampling rounds.\footnote{We used two separate functions of the form $f(t)=a+bt^\alpha$, where $t$ denotes the cumulative training sample size: one such function over the pre-switch period ($t<16,000$), and another over the post-switch period ($t\geq 16,000$).}

From this simple dynamic learning environment, we can draw some important insights. First, the severe under-performance of the pure schemes in times of mismatch indicate that single model-based active selections are strongly overfitted to the scoring model class. This pattern is consistent with findings in the aforementioned studies. This kind of mismatch even causes the \emph{naive\_adaptive} scheme (which is bound to choose a scorer from a single class in each round) to momentarily underperform passive sampling after the switch.\footnote{The \emph{margin-naive\_adaptive} scheme uses the \emph{logistic} model as scorer in every round prior to the switch, so its learning curve perfectly coincides with the learning curve of \emph{margin-logistic} in those rounds.} Seemingly, in our simulated experiment, active learning and model selection have closely intertwined objectives. The outperformance of the \emph{power} scheme over passing sampling shows that, even under dynamic (albeit deterministic) model transitions, active learning gains can be realized.

Secondly, we find no strong dominance among the competing schemes: the best performing strategy in any given round will yield suboptimal performance in some other rounds. Therefore, ranking the overall utility of these schemes hinges on defining preferences over the timing of the performance gains. In the extreme case where only performance in the final round mattered, \emph{kernel\_logistic} would clearly be the optimal scheme. In contrast, when gains are discounted over time, say geometrically at some rate $\rho\in(0,1)$, preponderance towards the logistic model may be warranted (e.g., the \emph{power} scheme, which is the optimal scheme in this simulation for the special case $\rho=1$, overweighs the logistic model in the initial rounds). Of course, a deterministic model switch is an oversimplification of reality. The design of the optimal scheme should evidently be sensitive to the \emph{distribution} of timing of model switches. The challenge of designing such a scheme in practice is compounded by our high uncertainty about the applicable discount rate (or whether geometric discounting applies at all) or the parameters of the stochastic process which drives model transitions. If our simple simulated experiment is any coarse approximation of reality, it suggests that ensemble scoring with time-varying weights may be a promising first step in addressing the practical challenge of model switches.

\subsection{Shifting Labels}\label{label_switch_subsection}
Label revisions for natural language processing (NLP) tasks occur from time to time in the real world. This holds especially true for segmentation, where the notion of ground truth may be more difficult to define than for downstream tasks such as part-of-speech tagging or dependency parsing. Whether coarser or more fine-grained segmentation labels is more useful hinges on the client's modeling goals. It is possible that, at some point during the training data creation, changes in client requirements necessitate revisions to the segmentation rules. This observation is not restricted to NLP tasks: such revisions are possible on any task where the ground truth may be somewhat subjective.

Our live experiment showed that major shifts in label distributions can significantly impact the selections made by active learning. This impact can be seen from the domain composition of the \emph{active} samples. In the live experiment, the candidate pool of sentences was sourced from two domains, which we shall refer to as domains ``A'' and ``B''. The two domains were equally represented in the candidate pool. In the actual $t=3$ batch---a batch which had an abnormally high proportion of domain “A” examples---domain A was overrepresented 3:1 relative to domain B. Had the labels been revised just prior to $t=3$, domain A would have been represented 1:2 against domain B, which would have been consistent with the proportions seen in the other pre-revision batches.

The robustness of active samples to label noise has been treated in the context of rater noise, e.g., in \citet{Sheng:2008:GLI:1401890.1401965}, \citet{Donmez:2009:ELA:1557019.1557053} and \citet{Mozafari:2014:SUC:2735471.2735474}. We note that the issue of shifting label distributions is orthogonal to the multiplicity of raters with heterogeneous skills. In fact, the issue can exist even in the case of a single high-skilled rater and this type of label noise may not be diversified away by using more raters or by improving their skills. 

%The reader will be quick to point that the obsolescence of labels would be irrelevant for active learning if the cost of revising an example’s label were the same as the cost of fully re-annotating that example. However, on the Thai segmentation task, the revision cost turned out to be substantially lower than the full annotation cost. Moreover, revised labels and new labels are not perfect substitutes: labels that have been subject to one or several revisions tend to be of higher quality than new labels. In the case of streaming data, gathering enough examples to make up for the discarded data could take a substantially long time. In such settings, simply throwing out mislabeled data may come with significant time cost. In multi-layer annotation settings, i.e., when each example receives a vector of labels (each label being associated with a single classification task), discarding a labeled example amounts to losing the whole vector of labels.

To the best of our knowledge, we have yet to find effective active learning designs that are robust to the potential of large shifts in the label distribution. However, we do note that these risks are, to a certain extent, naturally mitigated in systems in which the data are imposed expiration limits. Such limits, for example, can arise from maintenance or resource constraints. A six-month limit would imply that, six months after the most recent major label revision, any active example retained would have been selected per the correct labels. Such a limit provides a deterministic bound on the duration of the impact of incorrect labeling. Hard limits may however result in sharp performance kink, e.g., if large chunks of data get discarded altogether within a short time frame, or even in the phasing out of the whole dataset if the data is collected only sporadically. More gradual limits may be desirable to avoid drastic shifts in the training data distribution.

\subsection{Measuring Performance}\label{measure_subsection}
Accurate evaluation of learning curves is critical for making decisions, such as choosing the best scheme for deployment among competing alternatives, or stopping annotation once a performance target has been met. However, sufficient accuracy requires repeated measurements of the learning curve, which can be prohibitively costly to obtain in a live setting. Another challenge arises when the incremental labeled sample is small relative to the existing labeled pool: too small performance deltas can result in low-powered comparisons. Even deterministic selection schemes (e.g., margin ranking of a static unlabeled pool) may rely on an initial model trained on a seed labeled set. Such a seed set is a nuisance parameter which should be ``integrated out'' by running the comparison using different randomly selected seed sets. However, in a single-run live evaluation, we do not get the benefit of inference based on multiple alternative seeds. The passive learning curve itself is subject to sampling variability. This can make the comparison of active versus passive sampling particularly noisy.

The small sample issue that is inherent to the estimation of live learning curves may be addressed in two ways. In the first proposed approach, a parametric model of the learning curve is to be estimated via Bayesian inference using informative priors elicited from theoretical guarantees (e.g., expected convergence rate) or from actual performance measurements taken in related settings (e.g., other live experiments). The second approach is given by \citet{bheavlin-JSM}, who showed that the true performance variance of a trained model can be approximated using the ``half-sampling'' variance, i.e., the  performance variance of instances of the model trained on different half subsets of the training set.

\subsection{Scoring Uncertainty}\label{scoring_uncertainty_subsection}
Certain classes of models, notably neural networks, can be subject to large training variability. When these models are used as scorers, as was done in our study, their training variability can induce variability among the active selections. On different training runs of the DNN segmenter that used the same hyperparameter values, we observed meaningful differences among the proposed active batches. For example, out of ten runs at $t=13$, the average sentence length on domain B in a proposed batch ranged from 11.6 to 14.5 tokens, while the proportion of domain B sentences varied from 50\% to 76.8\%. The training-based variability of the active samples---another source of noise in the assessment of sampling gain---cannot be integrated out from a single-run live measurement. Due to the path dependency of active sampling, the impact of this training variability compounds over the sampling iterations and may become substantial.

As a workaround, our experiment used the following strategy: to approximate the best DNN segmenter (which is a deterministic quantity) and therefore reduce the variance of the scorer, we trained ten separate DNN models at each round, and selected the best performing model (based on validation F1 score) as scoring model. Even fairly good approximations (e.g., based on ten independent training runs) may significantly reduce the scoring variance compared to a single model.

\section{Conclusion}\label{conclusion}
Active learning strategies can greatly reduce the data acquisition cost of building supervised models. We put such a strategy--margin sampling--to test in the context of a real-time data collection task, and found substantial performance gains from  active sampling over the baseline strategy of passive sampling. However, our experiment also emphasized several practical challenges, from model and labeling uncertainty to measurement noise, that have a bearing on the usefulness of existing active learning solutions. While past research has brought some important insights on these issues, the solutions that have been proposed only partially address the complexity of live environments which our study has exemplified. We have outlined some tentative directions for tackling each of these challenges. Beyond our proposals, the design of active learning algorithms that are robust to these and other unforeseen challenges remains a vastly open area for further research.

\bibliography{pygmalion_workshop}
\bibliographystyle{icml2019.bst}

\end{document}